\UseRawInputEncoding
\pdfoutput=1
\documentclass[acmsmall]{acmart}
\AtBeginDocument{%
  }

\received{2024-10-31}
\received[accepted]{2025-03-31}

\acmConference[ISSTA' 25]{The 34th ACM SIGSOFT International Symposium on Software Testing and Analysis }{June 25--28, 2025}{Trondheim, Norway}

\acmISBN{2994-970X/2025/7 - ISSTA035}

\usepackage{bbm}
\usepackage{enumitem}
\usepackage{algorithm}
\usepackage{algorithmicx}
\usepackage[noend]{algpseudocode}
\usepackage{multirow}
\usepackage{pifont}
\usepackage{graphicx}
\graphicspath{{}}
\usepackage{subcaption}
\usepackage{tcolorbox}

\newcommand{\thx}[1]{\textcolor{black}{#1}}

\usepackage{xspace}
\newcommand{\name}{\texttt{FADE}\xspace}
\newcommand{\nameb}{\texttt{FADE}$_\texttt{r}$\xspace}

\begin{document}
\setcopyright{acmlicensed}
\acmYear{2025}
\acmVolume{2}
\acmNumber{ISSTA}
\acmArticle{ISSTA035}
\acmMonth{7}
\acmDOI{10.1145/3728910}
\title{Testing the Fault-Tolerance of Multi-sensor Fusion Perception in Autonomous Driving Systems}
\author{Haoxiang Tian}\authornote{This work was done at Nanyang Technological University as a visiting student.}
\email{tianhaoxiang20@otcaix.iscas.ac.cn}\authornotemark[3]
\affiliation{
  \institution{Institute of Software Chinese Academy of Sciences}\authornote{Affiliated with Key Lab of System Software at CAS, State Key Lab of Computer Science at Institute of Software at CAS, University of CAS, Beijing, China. CAS is the abbreviation of Chinese Academy of Sciences}
  \country{China}
}
\affiliation{
  \institution{Nanyang Technological University}
  \country{Singapore}
}

\author{Wenqiang Ding}
\email{dingwenqiang23@otcaix.iscas.ac.cn}
\affiliation{
  \institution{Institute of Software Chinese Academy of Sciences}
  \country{China}
}
\affiliation{
  \institution{University of CAS, Nanjing, Nanjing Institute of Software Technology}
  \country{China}
}

\author{Xingshuo Han}
\email{xingshuo001@e.ntu.edu.sg}
\affiliation{
  \institution{Continental-NTU Corporate Lab, Nanyang Technological University}
  \country{Singapore}
}

\author{Guoquan Wu}
\email{gqwu@otcaix.iscas.ac.cn}\authornotemark[3]\authornote{Guoquan Wu and Jun Wei are the corresponding authors}
\affiliation{
  \institution{Institute of Software Chinese Academy of Sciences}
  \country{China}
}
\affiliation{
  \institution{Nanjing Institute of Software Technology, University of CAS, Nanjing}
  \country{China}
}

\author{An Guo}
\email{guoan218@smail.nju.edu.cn}
\affiliation{
  \institution{Nanyang Technological University}
  \country{Singapore}
}
\affiliation{
  \institution{Nanjing University}
\country{China}
}

\author{Junqi Zhang}
\email{jqzh@ustc.edu.cn}
\affiliation{
  \institution{University of Science and Technology of China}
  \country{China}
}

\author{Wei Chen}
\email{chenwei@otcaix.iscas.ac.cn}\authornotemark[3]
\affiliation{
  \institution{Institute of Software Chinese Academy of Sciences}
  \country{China}
}

\author{Jun Wei}
\email{wj@otcaix.iscas.ac.cn}\authornotemark[3]\authornotemark[2]
\affiliation{
  \institution{Institute of Software Chinese Academy of Sciences}
  \country{China}
}

\author{Tianwei Zhang}
\email{tianwei.zhang@ntu.edu.sg}
\affiliation{
  \institution{Nanyang Technological University}
  \country{Singapore}
}

\renewcommand{\shortauthors}{Haoxiang Tian et al.}

\begin{abstract}
  Production-level Autonomous Driving Systems (ADSs), such as Google Waymo~\cite{waymo} and Baidu Apollo~\cite{apolloauto}, typically rely on the multi-sensor fusion (MSF) strategy to perceive their surroundings. This strategy increases the perception robustness by combining the respective strengths of the cameras and LiDAR, directly affecting the safety-critical driving decisions of autonomous vehicles (AVs). However, in real-world autonomous driving scenarios, both cameras and LiDAR are prone to various faults that can significantly impact the decision-making and subsequent behaviors of ADSs. It is important to thoroughly test the robustness of MSF during development. Existing testing methods only focus on the identification of corner cases that MSF fails to detect. However, there is still a lack of investigation on how sensor faults affect the system-level behaviors of ADSs. 
  
  To address this gap, we present \name, the \textbf{\textit{first}} testing methodology to comprehensively assess the fault tolerance of MSF perception-based ADSs. We systematically build fault models for both cameras and LiDAR in AVs and inject these faults into MSF-based ADSs to test their behaviors in various testing scenarios. To effectively and efficiently explore the parameter spaces of sensor fault models, we design a feedback-guided differential fuzzer to uncover safety violations of ADSs caused by the injected faults. We evaluate \name on Baidu Apollo, a representative and practical industrial ADS. The evaluation results demonstrate the practical values of \name, and disclose some useful findings. We further conduct physical experiments using a Baidu Apollo 6.0 EDU AV to validate these findings in real-world settings. 
\end{abstract}

\ccsdesc[500]{Software and its engineering~Software verification and validation}

\keywords{Autonomous Driving System, Fault Tolerance, Simulation Testing}


\maketitle

\section{Introduction}
In autonomous driving systems (ADSs), perception serves as a foundational module, as its output directly affects the safety-critical driving decisions for autonomous vehicles (AVs). Production-level ADSs, such as Google Waymo~\cite{waymo} and Baidu Apollo~\cite{apolloauto}, typically adopt a Multi-Sensor Fusion (MSF)-based perception strategy. It mainly leverages both cameras and LiDAR as the primary sensors to collect images and 3D point cloud data, respectively. These two modal data are processed separately using different deep learning models and subsequently fused to generate the final perception results. Compared to single-sensor perception, MSF improves the overall perception accuracy and increases the tolerance to sensor-specific errors, making ADSs more robust and reliable. However, in the highly complex and dynamic real-world driving environment, sensors are susceptible to various faults during vehicular operations. For example, the camera lens may become obstructed or damaged; the LiDAR may be misaligned due to the vehicle bump. These faults could compromise the quality of sensor data~\cite{bagloee2016autonomous, matos2024survey}. It remains unknown whether MSF-based ADS can still make safe decisions and actions in such situations. Therefore, it is crucial to test the fault tolerance of MSF-based ADSs under various types of sensor faults.

\thx{Existing works~\cite{zhong2022detecting, gao2024multitest, 84, gaobenchmark} primarily focus on generating corner cases of traffic environments to detect the errors of perception models, while overlooking the impacts of perception errors on system-level safety (e.g., decision-making, action control). A few works \cite{71, 72} test the effects of perception errors on AV crashes. However, they generate adversarial sensor inputs from scenarios rather than considering the inherent sensor faults. Secci \cite{camera_failure} and Ceccarelli \cite{duplicated} inject camera failures into the ADS to find safety violations. This approach only focuses on the effects of camera faults on the single-sensor (camera-only) ADS, without considering LiDAR faults and MSF in ADSs. One key advantage of MSF is its ability to compensate for the errors in a single sensor. Thus, this approach cannot accurately test and assess the fault tolerance of industrial-grade MSF-based ADS. Additionally, it does not guarantee that the detected safety violations are indeed caused by the injected sensor faults, as some of them may occur even without fault injection.}

\thx{To bridge this gap and test the system-level fault tolerance of MSF-based ADS against multi-sensor faults, we model real-world camera and LiDAR faults, and inject them into the ADS to identify their resulting safety violations.} However, there are two challenges to be addressed:

\begin{itemize}
    \item \textbf{Challenge 1: how to systematically and comprehensively model the sensor faults in real-world traffic for AVs.} The diverse and complex nature of traffic makes it difficult to comprehensively capture the unpredictable conditions affecting cameras and LiDAR on AVs.
    \item \textbf{Challenge 2: how to accurately identify the system-level safety violations caused by the injected sensor faults.} \thx{As the ADS is a highly coupled multi-component deep learning system, it is non-trivial to guarantee the discovered safety violations of the ADS indeed arise from the injected sensor faults.} 
\end{itemize}

In this paper, we propose \name, the \textbf{\textit{first}} \textbf{FA}ult-tolerance testing metho\textbf{D} to \textbf{E}valuate the multi-sensor perception of ADSs. \textbf{To address Challenge 1,} \name systematically \thx{categorizes} sensor faults as \textit{active faults} and \textit{passive faults}. It further subdivides these faults based on sensor components. Subsequently, \name builds the comprehensive fault models for the camera and LiDAR in real-world traffic. \textbf{To address Challenge 2,} 
\name designs and implements a differential fuzzer for sensor fault injection and system-level fault tolerance testing. This fuzzer evaluates the functional safety of the ADS under sensor faults and identifies their resulting safety violations. Our technical contributions are elaborated below. 

\textbf{1. Sensor Fault Modeling.} Specifically, \name \thx{categorizes} the sensor faults that possibly occur in real-traffic driving as active and passive ones, and utilizes FMEA \cite{Gilchrist1993ModellingFM, stamatis2003failure} to model them from the perspectives of sensor components and environmental factors.
\textit{Active faults} arise from the damage of sensors. For example, it could occur when the lens of the camera is damaged by external objects (e.g., stones or debris hitting the lens), resulting in cracks or partial visual obstruction. Similarly, an active fault in \thx{LiDAR enclosure} may occur when the AV is driving on a bumpy road, causing a shift in the sensor's mounting position. \textit{Passive faults} originate from the objects in the driving environment (e.g., weather, signals) that directly interfere with the normal operation of sensors. For example, the camera may get a passive fault caused by raindrops, snow grains, or mist on its lens. The LiDAR may have a passive fault in the \thx{processing unit} when the AV encounters a strong light source (such as a high beam), which makes its detector units occupied by strong light.

\textbf{2. Differential Fuzzer-based Sensor Fault Tolerance Testing.} The goal of \name is to test whether the MSF-based ADS is capable of functioning \thx{safely} in the presence of sensor faults during AV driving. 
To achieve this, \name designs and employs a genetic algorithm (GA)-guided differential fuzzer, to test the ADS with and without sensor faults, and identify its safety violations caused by injected sensor faults. 

We demonstrate the effectiveness of \name on a widely-used industrial L-4 ADS, Baidu Apollo \cite{apolloauto}, which adopts the MSF perception strategy \cite{baidu}. The results of simulation experiments show that \name can effectively and efficiently discover safety violations of Apollo caused by sensor faults. Furthermore, we conduct the \textbf{\textit{first}} physical experiments on multi-sensor faults in MSF-based AVs, to validate the authenticity and practical significance of our sensor fault models and the findings from simulation experiments. 
\thx{More than 60\% of our found safety violations of Apollo caused by injected sensor faults can be reproduced in physical experiments, which} demonstrates that our approach and findings hold substantial relevance for real-world AVs.

In summary, the paper makes the following contributions:

\begin{itemize}[leftmargin=*]
    \item \textbf{Originality.} To the best of our knowledge, we conduct the first exploration on the fault tolerance of MSF-based ADSs. Our findings can help understand how the system-level safety of MSF-based ADSs is affected by sensor faults in real-world traffic. 
    \item \textbf{Approach.} We propose \name, an automated sensor fault injection and sensor fault-tolerance testing approach for MSF-based ADSs. It systematically models sensor faults that AVs may encounter in real-world traffic, and employs a GA-guided differential fuzzer to identify the safety violations of ADSs caused by sensor faults.
    \item \textbf{Evaluation.} We evaluate \name on the representative industrial MSF-based ADS, Apollo. The results of our simulation experiments demonstrate that \name can effectively and efficiently discover safety violations of Apollo caused by injected sensor faults. The results of our physical experiments demonstrate the practical significance of our findings. 
\end{itemize}

\section{Background and Related Work}

\subsection{Perception in Autonomous Vehicles}
\label{sec:Cameras and LiDAR in Autonomous Vehicles}
In ADSs, the perception module relies on various sensors to detect and interpret the surrounding environment \cite{campbell2018sensor}. Cameras and LiDAR are the two critical sensors used for this purpose.
These sensors form the backbone of the perception module, enabling the AV to perceive the surrounding environment with high fidelity \cite{feng2021review, qian20223d}.

\textbf{Camera-based perception.}
A typical camera mainly comprises five components: lens, camera body, Bayer filter, image sensor, and image signal processor (ISP) \cite{phillips2018camera}. The lens plays a critical role in determining the image quality, focusing light onto the sensor, and enabling image reproduction \cite{altenburg2013lens}. The camera body houses and protects internal electronics, while also shielding sensitive parts from environmental exposure. The Bayer filter enables color capture by placing red, green, and blue filters over the sensor's pixels \cite{sabins2020remote, baumgartner2009benchmarks}. The image sensor converts captured light into electrical signals, forming the digital image. Finally, the ISP processes this data, enhancing image quality by applying various corrections and producing the final output image \cite{may2004fotografia}. 

\textbf{LiDAR-based Perception.}
A typical LiDAR \thx{has} five main components, including laser emitter, scanner, receiver, processing unit and LiDAR \thx{enclosure \cite{chen2017multi}}. The laser emitter generates laser pulses that are projected into the environment \cite{liu2016ssd}. These pulses reflect off surrounding objects and return to the receiver \cite{li2016vehicle}. The scanner orchestrates the laser's movement to cover a 360-degree field of view or specific sectors, enabling comprehensive spatial mapping. Finally, the processing unit computes the distance and shape of surrounding objects by measuring the time it takes for each pulse to return \cite{asvadi20163d}. LiDAR enclosure is an accessory used to protect the \thx{LiDAR} lines. Together, these components produce high-resolution, 3D point clouds that enhance the depth perception and allow the system to detect and interpret the vehicle’s environment with precision \cite{yin2021center}.

\textbf{Multi-Sensor Fusion-based Perception.} 
Cameras capture high-resolution visual data, providing contextual information such as road signs, object appearances and motion changes. This visual input allows for object classification and recognition, which are essential for safe navigation \cite{rosique2019systematic}. However, the 2D camera imaging lacks depth information of the 3D driving spaces. LiDAR, on the other hand, uses laser beams to measure distances and generate high-precision 3D point clouds of the traffic environment \cite{yang2018pixor}. This sensor is highly effective in accurately identifying the position and shape of objects, including other vehicles, pedestrians, and obstacles. \thx{LiDAR performs well in diverse lighting and weather conditions, which complements the weaknesses of cameras. However, LiDAR struggles to capture detailed texture information (e.g., color), which can be provided by cameras \cite{frossard2018end}.} By integrating LiDAR’s depth data with the texture details from cameras, MSF algorithms can enhance object detection beyond the capabilities of either sensor used alone \cite{liang2018deep,chen2017multi}.

\subsection{Perception Testing of ADSs}

\thx{The reliability of the perception module is critical for the safety and functionality of ADSs \cite{koopman2017autonomous}, making its testing essential. Existing works focus on two main aspects: (1) testing the perception models and (2) testing the impact of perception errors or sensor failures on ADSs.}

\subsubsection{\thx{Testing Perception Models}}
\thx{Many studies generate adversarial examples or corner cases for the perception models to identify their errors in scenario understanding (e.g., object detection and tracking). These approaches can be categorized into three types based on the perception strategies}

\textit{\thx{(1) Camera-based perception model testing}}.
\thx{(\romannumeral1) Some works \cite{61,64,65,66,67,80,81,88} generate adversarial perturbations or patches to mislead deep learning-based camera models (e.g., Faster R-CNN, YOLO), particularly targeting real-world objects like traffic signs. (\romannumeral2) Several works \cite{82,83,85,86} leverage GAN-based perceptual adversarial networks to deceive camera-based perception models. (\romannumeral3) Additionally, adversarial camouflage patterns \cite{63} are proposed to conceal 3D objects from detection. A metamorphic testing approach \cite{100} is designed to identify inconsistencies in obstacle detection.}

\textit{\thx{(2) LiDAR-based perception model testing.}}
\thx{(\romannumeral1) Some studies \cite{74, 77, christian2023generating, 101} generate 3D adversarial point clouds to attack LiDAR-based perception models. For example, Zhou et al. \cite{101} employ a metamorphic testing approach combined with fuzzing to detect errors in LiDAR obstacle perception. (\romannumeral2) Some works \cite{76, 78} identify the impact of critical adversarial locations, and use simple objectives or arbitrary reflective objects to fool LiDAR perception models. (\romannumeral3) A few works leverage occlusion patterns \cite{73}, or affine and weather transformations \cite{guo2022lirtest} to generate adversarial inputs to augment LiDAR point clouds. (\romannumeral4) Moreover, Wang et al. \cite{wang2021advsim} and Li et al \cite{79} use polynomial perturbations on trajectories of NPC vehicles and pedestrians, to test the LiDAR's ability to recognize adversarial dynamic objects and behaviors.}

\textit{\thx{(3) MSF-based perception model testing.}}
\thx{Zhong et al. \cite{zhong2022detecting} identify fusion errors cased by incorrect multi-sensor data integration. Xiong et al. \cite{84} generate adversarial samples by separately perturbing camera and LiDAR inputs while maintaining data correlation. Gao et al. \cite{gao2024multitest} synthesize real-world data and seek to insert objects into scenarios to uncover perception errors in MSF-based modules, evaluating the perception accuracy under challenging scenarios generated by the metamorphic testing approach. Meanwhile, Gao et al. \cite{gaobenchmark} summarize and implement a range of real-world corruption patterns on the MSF perception module, and test their impacts on the perception results.} 


\thx{However, these approaches primarily focus on generating adversarial examples or corner cases that exploit vulnerabilities in single-modal (camera-based or LiDAR-based) or MSF-based perception models. They do not assess how perception errors propagate to subsequent modules or impact the overall behavior of the ADS.}

\subsubsection{\thx{Testing the Effects of Perception Errors or Sensor Failures on ADSs}}

\thx{A few efforts have examined the effects of perception errors on the system-level consequences of ADSs. Cao et al. \cite{71} categorize different LiDAR spoofing attack patterns from previous studies to assess their impact beyond the perception stage and analyze their influence on the decision-making of ADSs. Additionally, Cao et al. \cite{72} manipulate the shape of 3D meshes by altering vertex positions, synthesizing point clouds and camera images to mislead the ADS into failing to detect objects, ultimately causing crashes.} \thx{However, these works primarily generate adversarial perception inputs from scenarios to induce the ADS's errors, rather than considering sensor faults that arise in real-world traffic.} 

\thx{Secci \cite{camera_failure} and Ceccarelli \cite{duplicated} inject camera failures into the ADS and test its behaviors to find safety violations.} \thx{However, this approach has two key limitations: (1) it only targets camera-based ADS, disregarding the role of LiDAR and MSF in mitigating camera faults for ADSs; (2) It does not verify whether the identified safety violations are genuinely caused by injected faults. Therefore, it fails to comprehensively and accurately evaluate the fault tolerance of MSF-based ADSs.}

\thx{Different from existing works, our study is the \textbf{first} to systematically test the system-level fault tolerance of MSF-based ADSs, identifying behavioral safety violations caused by various sensor faults. Our goal is to evaluate whether MSF-based ADSs are robust enough to maintain AV safety when encountering real-world sensor faults during driving.}

\section{Approach}
\begin{figure}[h]
  \centering
  \includegraphics[scale=0.29]{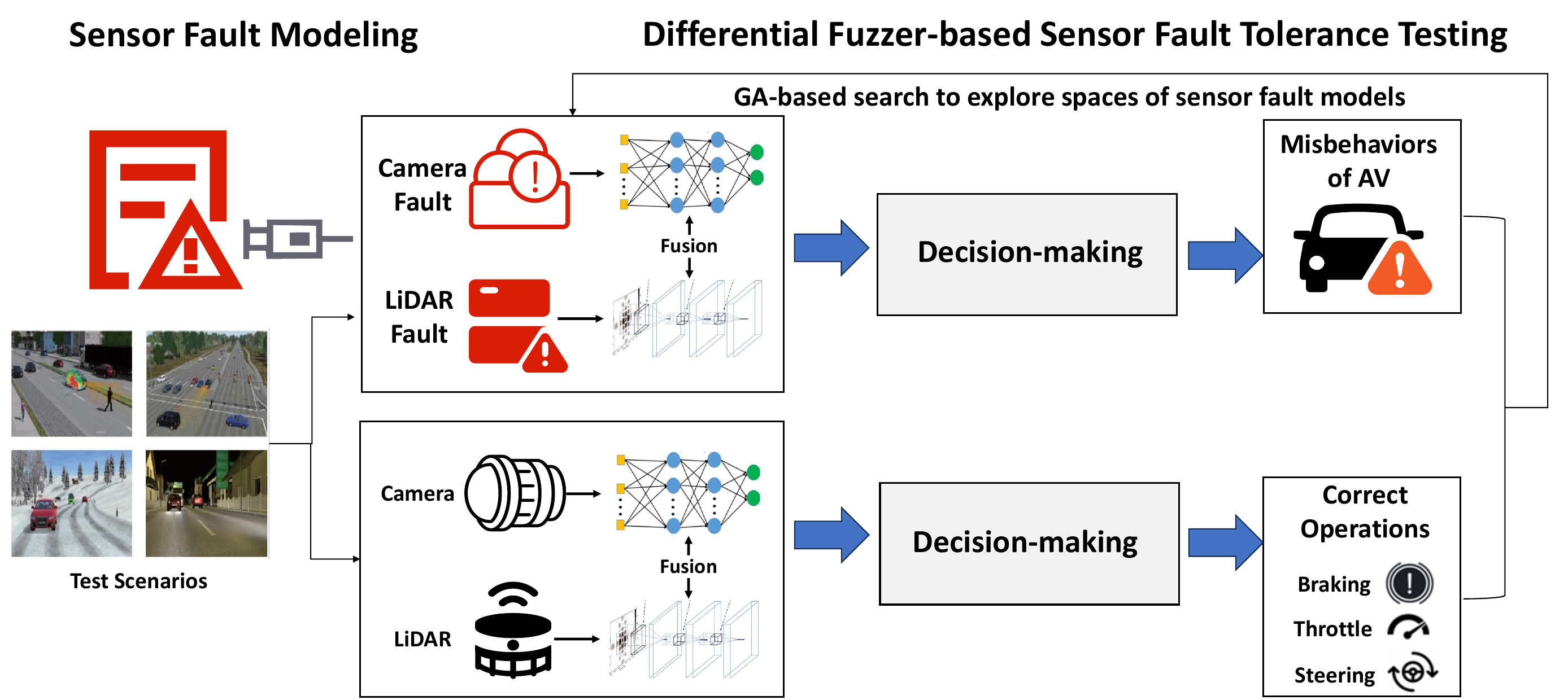}
  \caption{The overview of \name.}
  \label{overview}
  \end{figure}

\thx{Our objective is to test the fault-tolerance of MSF-based ADSs under camera and LiDAR faults that may occur during AV driving.} To this end, we introduce a novel approach: \name. Its overview is presented in Figure~\ref{overview}, which consists of two parts. \textbf{\ding{182} Sensor fault modeling}: \name systematically models faults that cameras and LiDARs may encounter in real-world traffic environments, including active faults and passive faults caused by various environment factors. \textbf{\ding{183} Differential Fuzzer-based Sensor Fault Tolerance Testing}: \name designs and implements a GA-guided differential fuzzer, which uses differential testing to test the performances of the ADS with and without sensor faults, and explores the space of fault models by a GA-based search to discover safety violations of the ADS caused by the injected sensor faults. Below we give details of each component.

\subsection{Sensor Fault Modeling}

Sensors in real-world driving environments may encounter various issues that affect their accuracy and reliability. These issues can be broadly categorized into two main types: active faults and passive faults. Active faults arise from internal sensor malfunctions or damages, directly disabling or degrading their functionality. These faults typically result from component failures or wear. In contrast, passive faults stem from external environmental factors (e.g., an obstacle), rather than sensor damage or malfunction. These environmental factors can degrade the sensor’s ability to capture or interpret perception data accurately.

To systematically model these sensor faults in traffic scenarios, we classify them from the perspectives of sensor components and environmental factors. Further, we develop fault models to simulate possible failures in cameras and LiDARs under real-world AV driving conditions. These models enable the construction of a comprehensive fault model library for AV perception, facilitating realistic fault injection experiments. A summary of camera and LiDAR fault-injection models is provided in Table~\ref{camera} and Table~\ref{lidar}.

\subsubsection{Camera Fault Model}
The camera fault models are categorized into 16 types, including 7 types of active faults and 9 types of passive faults. Due to page limits, we mainly describe one fault model and others are available at \textit{sensor\_fault\_models.pdf} in \url{https://github.com/ADStesting-test/FADE}.

\begin{table}[h]
\centering
\caption{Categorization of camera fault-injection models}
\footnotesize
\begin{tabular}{|c|c|c|c|}
\hline
\textbf{\begin{tabular}[c]{@{}c@{}}Fault\\ Category\end{tabular}} & \textbf{\begin{tabular}[c]{@{}c@{}}Fault\\ Type\end{tabular}}   & \textbf{An Example in Real Traffic}                                                                                                                                        & \textbf{\begin{tabular}[c]{@{}c@{}}Faulty\\ Componet\end{tabular}}              \\ \hline
\multirow{7}{*}{\rotatebox[origin=c]{90}{\textbf{Active \quad Fault}}}                            & Deflection                                                      & Due to bumpy roads, the camera deflects during AV driving                                                                                                                  & \multirow{3}{*}{\begin{tabular}[c]{@{}c@{}}Camera\\ Body\end{tabular}}          \\ \cline{2-3}
                                                                  & Displacement                                                    & Due to bumpy roads, the camera is misaligned during AV driving                                                                                                             &                                                                                 \\ \cline{2-3}
                                                                  & \begin{tabular}[c]{@{}c@{}}Internal\\ Dirt\end{tabular}         & \begin{tabular}[c]{@{}c@{}}The dirt accumulation inside caused by\\driving outside or changing temperature for a long time\end{tabular}                    &                                                                                 \\ \cline{2-4} 
                                                                  & Broken Lens                                                      & The splashing foreign object(e.g.,sand, gravel) hits the lens                                                                                                               & \multirow{2}{*}{Len}                                                            \\ \cline{2-3}
                                                                  & \begin{tabular}[c]{@{}c@{}}Lens Brightness\\ Change\end{tabular} & \begin{tabular}[c]{@{}c@{}}Long-term usage in high-temperature or seaside cause brightness\\ issues with basic components(e.g.,shutter,diaphragm,iris) of len\end{tabular} &                                                                                 \\ \cline{2-4} 
                                                                  & Blur                                                            & The blur introduced by malfunction of complex inside circuit                                                                                                               & \multirow{2}{*}{\begin{tabular}[c]{@{}c@{}}Image Sen-\\ sor\& ISP\end{tabular}} \\ \cline{2-3}
                                                                  & Internal Scatter                                                & The color noise caused by malfunction of image signal processor                                                                                                            &                                                                                 \\ \hline
\multirow{9}{*}{\rotatebox[origin=c]{90}{\textbf{Passive \quad Fault}}}                           & Lens Occlusion                                                   & The lens is covered with plastic bag or paper during AV driving                                                                                                            & \multirow{7}{*}{Len}                                                            \\ \cline{2-3}
                                                                  & External Scatter                                                & The surface of the camera's lens is contaminated with mud spots                                                                                                             &                                                                                 \\ \cline{2-3}
                                                                  & Dust                                                            & The surface of the camera's lens is contaminated with dust                                                                                                                  &                                                                                 \\ \cline{2-3}
                                                                  & Raindrops                                                       & Raindrops appear on the lens along with rainlines during rainfall                                                                                                           &                                                                                 \\ \cline{2-3}
                                                                  & Snow Grains                                                     & The deposit of snow grains on the lens during snowfall                                                                                                                     &                                                                                 \\ \cline{2-3}
                                                                  & Mist                                                            & Fogging of lens caused by high humidity and temperature differences                                                                                                         &                                                                                 \\ \cline{2-3}
                                                                  & Ice                                                             & Temperature drops below zero degrees, resulting in ice on the lens                                                                                                          &                                                                                 \\ \cline{2-4} 
                                                                  & Overexposure                                                    & \begin{tabular}[c]{@{}c@{}}Under strong light sources such as high beams or reflective\\ surfaces, camera receives too much light\end{tabular}                             & \begin{tabular}[c]{@{}c@{}}Image\\ Sensor\end{tabular}                          \\ \cline{2-4} 
                                                                  & \begin{tabular}[c]{@{}c@{}}White Balance\\ Shift\end{tabular}   & \begin{tabular}[c]{@{}c@{}}At sunset, the camera image appears in red and orange tones\\ due to white balance shift\end{tabular}                                           & \begin{tabular}[c]{@{}c@{}}Bayer\\ Filter\end{tabular}                          \\ \hline
\end{tabular}
\label{camera}
\end{table}


\textbf{Raindrops on the lens.}
Raindrops and rainlines act as random streaks of water on the lens, causing scattering, refraction, and partial occlusion of the camera’s field of view. Under natural conditions the raindrops exhibit the tilted shape instead of the linear shape \cite{roser2009video}. Therefore, when simulating the effect of raindrops and rainlines, the angle and shape changes of raindrops are introduced. Meanwhile, to make the raindrop effect more realistic, the transparency (or brightness) of raindrops is also adjusted \cite{garg2006photorealistic}. The image $I_{r}$ captured by the camera with raindrops on the lens is:
\begin{equation}
    I_r (x, y) = (1 - L_r (x, y)) \cdot I_e (x, y) + L_r (x, y) \cdot (t_r \cdot I_e (x, y) + (1 - t_r) \cdot N_r (x, y))
\end{equation}
\begin{equation}
    t_r (x, y) \sim U (t_{min}, t_{max}), \quad N_r (x, y) \sim \mathcal N (0, \sigma_r)
\end{equation}
where $I_e (x, y)$ is the original pixel intensity of the image captured by the camera without raindrops on the lens. $N_r (x, y)$ is a random noise factor simulating the refraction-induced distortion. $t_r$ is a transparency factor, simulating the partial occlusion of the camera's view due to the rain streaks. $L_r (x, y)$ is the rain line mask, calculated as:
\begin{equation}
    L_r (x, y) = \Sigma_{i=1}^{n_r} \mathcal H \left( \frac{(y-y_i)-tan(\theta_r^i)(x-x_i)}{l_r^2} \right)
\end{equation}
\begin{equation}
    l_r \sim U (l_{min}, l_{max}), \quad \theta_r \sim U (\theta_{min}, \theta_{max})
\end{equation}
$L_r (x, y)$ represents the spatial distribution of the rain streaks on the lens. Each streak can be modeled as a linear segment on the image with the length $l_r$ and angle $\theta_r$. $(x_i, y_i)$ is the starting position of a rain line. $\mathcal H (f)$ is a Heaviside function (or step function) that returns 1 if $f$ is within the rainline length, and 0 otherwise, controlling the spatial extent of each rainline.

\subsubsection{LiDAR Fault Model}

The LiDAR fault models are categorized into 8 types, including 4 types of active faults and 4 types of passive faults. We introduce two LiDAR fault models and others are available at \textit{sensor\_fault\_models.pdf} in \url{https://github.com/ADStesting-test/FADE}.

\begin{table}[h]
\footnotesize
\centering
\caption{Categorization of LiDAR fault-injection models}
\begin{tabular}{|c|c|c|c|}
\hline
\textbf{\begin{tabular}[c]{@{}c@{}}Fault\\ Category\end{tabular}} & \textbf{\begin{tabular}[c]{@{}c@{}}Fault\\ Type\end{tabular}}          & \textbf{An Example in Real Traffic}                                                                                                                             & \textbf{\begin{tabular}[c]{@{}c@{}}Faulty\\ Component\end{tabular}}        \\ \hline
\multirow{4}{*}{\rotatebox[origin=c]{90}{\textbf{Active Fault}}}                            & Deflection                                                             & Due to bumpy roads, the orientation of LiDAR changes                                                                                           & \multirow{2}{*}{\begin{tabular}[c]{@{}c@{}}LiDAR\\ enclosure\end{tabular}} \\ \cline{2-3}
                                                                  & Displacement                                                           & Due to bumpy roads, the position of LiDAR is displaced                                                                                                     &                                                                            \\ \cline{2-4} 
                                                                  & Beam Loss                                                              & \begin{tabular}[c]{@{}c@{}}Due to long-term wear\&tear and aging,\\laser beam of LiDAR reduces \end{tabular}                                                                                       & Emitter                                                                    \\ \cline{2-4} 
                                                                  & Line Fault                                                             & The noise caused by malfunction of internal circuits                                                                                                            & \begin{tabular}[c]{@{}c@{}}Process-\\ ing Unit\end{tabular}                \\ \hline
\multirow{4}{*}{\rotatebox[origin=c]{90}{\textbf{Passive \quad Fault}}}                           & \begin{tabular}[c]{@{}c@{}}Electromagnetic\\ Interference\end{tabular} & \begin{tabular}[c]{@{}c@{}}When AV passes airports or power plants,these areas will\\ generate electromagnetic wave interference\end{tabular}                   & \multirow{2}{*}{Receiver}                                                  \\ \cline{2-3}
                                                                  & Crosstalk                                                              & \begin{tabular}[c]{@{}c@{}}NPC vehicles using LiDAR drive around AV,and their emitted\\signals cause confusion to receiving channel of AV's LiDAR\end{tabular} &                                                                            \\ \cline{2-4} 
                                                                  & \begin{tabular}[c]{@{}c@{}}Rain and Snow\\ Pollution\end{tabular}      & \begin{tabular}[c]{@{}c@{}}There are foreign objects (such as rain, snow, mist, mud)\\ covering LiDAR's surface,limiting the LiDAR's field of view\end{tabular}   & Scanner                                                                    \\ \cline{2-4} 
                                                                  & \begin{tabular}[c]{@{}c@{}}Strong Light\\ Interference\end{tabular}    & \begin{tabular}[c]{@{}c@{}}The measurement distance and point-cloud density are reduced\\ due to strong light occupying detector units\end{tabular}             & \begin{tabular}[c]{@{}c@{}}Process-\\ ing Unit\end{tabular}                \\ \hline
\end{tabular}
\label{lidar}
\end{table}

\textbf{Deflection of LiDAR enclosure.}
This fault is introduced by LiDAR's deflection of the vertical direction to the direct direction and scan angle direction. When the deflection of the vertical direction is perpendicular to the scan angle direction, the resulting vertical error will be negligible. However, when the deflection is parallel to the scan angel direction, the induced vertical error reaches the maximum value \cite{jekeli2019deflections}. The magnitudes of the rotations are defined as $\xi$ and $\eta$ components of the deflection of the vertical direction, and the resulting $R_G$ is formalized as:
\begin{equation}
    R_G = \left[ \begin{array}{ccc} cos ( \eta ) & sin ( \xi ) sin ( \eta ) & sin ( \xi ) cos ( \eta ) \\ 0 & cos ( \xi ) & - sin ( \xi ) \\ -sin ( \eta ) & cos ( \xi ) sin ( \eta ) & cos ( \xi ) cos ( \eta ) \end{array} \right]
\end{equation}

\textbf{Displacement of LiDAR enclosure.}
This fault model is defined by the grid mean approximation and triangular grid approximation, which have been applied experimentally to generate reference data for 3D data in the spatial domain \cite{kang2007computing, notbohm2013three, gawronek2019measurements}. The grid mean approximation method includes grid point errors and forms grids on the $x$ and $y$ planes based on irregularly distributed spatial data. Thereafter, the $z$ coordinates of the data in the grid are averaged to determine the representative point of each grid. The grid mean approximation method uses a multiple regression analysis technique and calculates the displacement of a structure using structural information such as strain, stress, displacement, and z-coordinates:
\begin{equation}
    P_j (x, y, z) = ( \varepsilon, Z_j), \quad Z_j = \frac{1}{n_j} \Sigma_{i=1}^{n_j} \Sigma_{j=1}^{m} Z_{ji}, \quad \varepsilon = N_1 \varepsilon_1 + N_2 \varepsilon_2 + N_3 \varepsilon_3
\end{equation}
where $Z_{ji}$ represents the $Z$ coordinate value of the $i$-th coordinate data included in the $j$-th space. The point $P_j$ is set as reference data in the center of the grid. The shape functions $N_1, N_2, N_3$ are calculated through natural coordinates. 

\subsection{Differential Fuzzer-based Sensor Fault Tolerance Testing}
\thx{Based on the fault models of camera and LiDAR, \name injects sensor faults into the MSF-based ADS, and employs a differential fuzzer to assess its fault tolerance and identify the safety violations in the ego vehicle. 
Our sensor fault-tolerance testing procedure is detailed in Algorithm \ref{al::differential}. The explanations of notations used in this algorithm are given as Table~\ref{notations}.}

\begin{table}[h]
\caption{Explanations of notations used in Algorithm \ref{al::differential}}
\small
\begin{tabular}{|c|c|c|c|}
\hline
\textbf{Notation}       & \textbf{Explanation}                              & \textbf{Notation}                                    & \textbf{Explanation}                                 \\ \hline
\textit{\textbf{$ts$}}    & a test scenario                                   & \textit{\textbf{$AD_o$}}                              & ego vehicle without fault                            \\ \hline
\textit{\textbf{$sf$}}    & a sensor fault                                    & \textit{\textbf{$rs^{ts}_o$}}      & execution result of $AD_o$ in $ts$                      \\ \hline
\textit{\textbf{$fp$}}    & an instance of sf injected into $AD_o$             & \multirow{2}{*}{\textit{\textbf{$RS^{ts}_{fp}$}}} & \multirow{2}{*}{\begin{tabular}[c]{@{}c@{}}differential testing result\\of $AD_f$ and $AD_o$ in $ts$\end{tabular}} \\ \cline{1-2}
\textit{\textbf{$AD_f$}} & ego vehicle with the injected $fp$          & &  \\ \hline    
\textit{SVF($RS^{ts}_{fp}$)} & \begin{tabular}[c]{@{}c@{}}assertion for safety violation\\caused by $fp$ in $ts$\end{tabular} & \textit{\textbf{$\Phi_o(rs^{ts}_o)$}}                & \begin{tabular}[c]{@{}c@{}}assertion for no safety violation\\ of $AD_o$ in $ts$\end{tabular}  \\ \hline
\end{tabular}
\label{notations}
\end{table}

Specifically, \name first generates test scenarios $\mathbb{TS}$, and instances ($\mathbb{FP}$) of the injected sensor fault $sf$. It then injects the sensor fault instance $fp$ into the ADS, and performs differential testing by comparing the behaviors of the ego vehicle with $fp$ ($AD_f$), and without $fp$ ($AD_o$) in the same test scenario $ts$. Safety violations caused by $fp$ are identified by analyzing the system-level performance differences between $AD_f$ and $AD_o$. Additionally, for each $sf$ in $ts$, \name leverages a multi-objective genetic algorithm to optimize its injected instance $fp$, further exploring its potential to trigger safety violations in the ADS.


\begin{algorithm}[h]
  \begin{small}
  \caption{Sensor fault-tolerance testing} 
  \label{alg::testing}
  \begin{algorithmic}[1]
    \Require
      Camera fault models $\mathbb{CFM}$, LiDAR fault models $\mathbb{LFM}$
    \Ensure
      $\mathbb{SV}$: Safety violations of ego vehicle caused by injected sensor faults
    \State{\thx{$\mathbb{SV}$ $\leftarrow \emptyset$, \textit{terminate} $\leftarrow False$}}
    \State{\thx{$\mathbb{TS} \leftarrow generate\_scenarios(num)$}} \textcolor{gray}{\Comment{generate a set of test scenarios}}
    \State{\thx{$\mathbb{SF} \leftarrow \mathbb{CFM} \bigcup \mathbb{LFM}$, $\mathbb{ISF} \leftarrow$ \textit{combination}($\mathbb{CFM}$, $\mathbb{LFM}$)}}
    \For{\thx{$isf \in \mathbb{ISF}$}}
    \If{\thx{$\forall i,j \in isf, i.pre = j.pre$}}
    \State{\thx{$\mathbb{SF} \leftarrow \mathbb{SF} \bigcup isf$}} \textcolor{gray}{\Comment{determine the sensor faults to be injected into the ADS}}
    \EndIf
    \EndFor
    \For{\thx{$\forall sf \in \mathbb{SF}$}}
    \For{\thx{$\forall ts \in \mathbb{TS}$}}
    \State{\thx{$rs^{ts}_o \leftarrow execute(ts, AD_o)$}}
    \If{\thx{$\Phi_o(rs^{ts}_o)$}}
    \State{\thx{$\mathbb{FP} \leftarrow initialization(sf, k)$}}\textcolor{gray}{\Comment{create initial instances of the injected sensor fault $sf$}}
    \For{\thx{$\forall fp \in \mathbb{FP}$}}
    \State{\thx{$AD_f \leftarrow$ \textit{inject\_faults}($fp, AD_o$)}}\textcolor{gray}{\Comment{inject the instance of $sf$ into the ADS}}
    \State{\thx{$RS_{fp}^{ts}$ = \textit{differential\_testing}$(AD_f, AD_o, ts)$}}\textcolor{gray}{\Comment{perform differential testing}}
    \EndFor
    \State{\thx{$\mathbb{RS} \leftarrow$ \textit{differential\_fuzzer}($sf, ts, \mathbb{FP}$)}}\textcolor{gray}{\Comment{optimize instances of $sf$ to find safety violation of $AD_f$}}
    \For{\thx{$RS_{fp}^{ts} \in \mathbb{RS}$}} \textcolor{gray}{\Comment{compare the results to identify ADS's safety violations caused by $sf$}}
    \State{\thx{$monitor \leftarrow SVF(RS_{fp}^{ts})$}}
    \If{\thx{$monitor$}}
    \State{\thx{$\mathbb{SV} \leftarrow \mathbb{SV}\bigcup RS_{fp}^{ts}$}}
    \EndIf
    \EndFor
    \EndIf
    \EndFor
    \EndFor
    \State {return $\mathbb{SV}$}
  \end{algorithmic}
  \label{al::differential}
\end{small}
\end{algorithm}
\thx{Considering that the simultaneous faults of multiple sensors are rare in real-traffic driving, \name primarily injects each single-sensor fault into the ADS. However, to balance practicality and testing thoroughness, \name also considers sensor co-faults, which refer to the multiple faults arising from the same environmental conditions. For example, both the \textit{overexposure} of camera and \textit{strong light interference} of LiDAR, arise from the strong light in the driving environment. Thus the two faults constitute a sensor co-fault, and \name injects the two faults simultaneously into the ADS.}

\subsubsection{\thx{Generating Test Scenarios}}
\thx{To evaluate the ADS's performance under sensor faults, \name generates test scenarios that reflect diverse real-world traffic conditions, leveraging naturalistic driving data \cite{feng2020testing1,feng2020testing2}. These scenarios are parameterized by variables including road topology, traffic density, and the behaviors of dynamic participants  (e.g., actions, speeds, accelerations, and relative positions of NPC vehicles and pedestrians). The scenarios are then generated through a constraint-based sampling of parameters within the Operational Design Domain (ODD) \cite{czarnecki2018operational}.
To ensure the validity and relevance of the generated scenarios, \name enforces constraints to prevent unrealistic cases that violate the ODD and invalid cases that lack meaningful interactions with the ego vehicle. Specifically, \name imposes the following constraints}:

\begin{enumerate}
\item $\forall i\in k, p_i(0) - p_{ego}(0) \le d_{max}$. $p_i(0)$ represents the position of the $i$-th participant at the initial time, and $k$ is the total number of participants in the scenario. $d_{max}$ represents the maximal distance of the sensors on the ego vehicle. This constraint requires the initial positions of participants not to be too distant from the ego vehicle’s initial position, ensuring the participants in the test scenario are more likely to enter the sensing range of the ego vehicle's sensors. 

\item $\forall i\in k, \exists t\in F \bigwedge rs \in R, p_i(t) \in rs \bigwedge p_{ego}(t) \in rs$. $F$ represents the total time steps of the test scenario execution, and $R$ represents the road map of the test scenario consisting of several road segments (represented by $rs$). This requires that each participant's trajectory has a same road segment as the ego vehicle's trajectory. Constraint (1) and constraint (2) ensure the ego vehicle has interactions with participants in the test scenario.

\item $\forall i\in k, t\in F, (v_{t+1}-v_t \le a_{max} \bigwedge v_i(t) \le v_{max}) \bigwedge (d_{i}(t) = dir_{rs}, p_i(t) \in rs)$. $v_i(t)$ represents the speed of participant $i$ at time $t$ and $v_{max}$ represents the speed limit of the road. $a_{max}$ represents the maximal acceleration of the participant (vehicle or pedestrian). $d_i(t)$ represents the direction of participant $i$ at time $t$ and $dir_{rs}$ represents the direction of the road segment $rs$ where $p_i(t)$ is located (for each pedestrian, $dir_{s}$ includes all directions). This constraint requires the speeds and directions of participants not to violate the realistic traffic dynamics during their motions.
\end{enumerate}



\subsubsection{\thx{GA-based Differential Fuzzer}}

\thx{For each sensor fault and co-fault $sf$ in a generated test scenario, \name initializes $k$ instances of $sf$ randomly and injects each fault instance $fp$ into the ADS to create $AD_f$. It then performs differential testing between $AD_f$ and $AD_o$ in the same scenario $ts$.} 
\thx{Using the recorded differential testing results, \name employs a multi-objective GA \cite{tian2022mosat} to optimize the instances of $sf$ to expose ADS's safety violations caused by $fp$ in $ts$.}
\noindent\thx{\textbf{Individual Encoding and Representation}. In $ts$, each instance of the injected sensor fault is encoded as an individual, consisting of one or two chromosomes. \name encodes each injected single sensor fault as a chromosome and each co-fault as two chromosomes. Each chromosome consists of a series of genes, and one gene corresponds to a parameter of the sensor fault model.}

\thx{The scores of each individual are calculated by a \textbf{multi-objective fitness function}. The optimal individuals are selected as parents, by building improved Pareto-optimal solutions from the individuals of the current and previous generations. The individuals of the next generation are generated by \textbf{variation operators} on parents. When the selected top $k$ excellent individuals remain the same in three consecutive generations, the optimization of instances for $sf$ in $ts$ terminates.}


\noindent\thx{\textbf{Multi-objective Fitness Function.}}
\thx{The fitness function optimizes two objectives: the additional risk introduced by $AD_f$ compared to $AD_o$, and the motion deviation between them. Based on these objectives, \name constructs successfully-improved Pareto-optimal solutions. Specifically, $fp$ represents an instance of the injected sensor fault $sf$.}

\thx{(1) Objective 1: the additional risk introduced by $AD_f$ compared to $AD_o$. The additional risk introduced by $fp$ during the driving of $AD_f$ compared to $AD_o$ in the same test scenario $ts$, is defined as $I(fp, ts)$:}
\thx{
\begin{equation}
    I(fp, ts) = METTC(AD_o, ts) - METTC(AD_f, ts)
\end{equation}}
\thx{METTC is the minimal estimated time for collision \cite{ettc}, which is widely used to measure the risk of the AV during driving \cite{ettc_a, ettc_point}. \thx{We} define an improved calculation of METTC as follows:}
\thx{
\begin{small}
\begin{equation}
METTC(AD_f, ts)=\min_{\substack{i \in [1,k] \\ t \in (0,F)}} \left(\frac{d.x(b_t^{AD_f}, b_t^{p_i})}{v.x_t^{p_i}-v.x_t^{AD_f}}+\frac{v.x_t^{p_i}-v.x_t^{AD_f}}{ac.x_t^{p_i}-ac.x_t^{AD_f}}, \frac{d.y(b_t^{AD_f}, b_t^{p_i})}{v.y_t^{p_i}-v.y_t^{AD_f}}+\frac{v.y_t^{p_i}-v.y_t^{AD_f}}{ac.y_t^{p_i}-ac.y_t^{AD_f}}\right)
\end{equation}
\end{small}}
\thx{where $p_i$ is the $i$-th participant in the scenario $ts$. $d.x(b_t^{AD_f}, b_t^{p_i})$ and $d.y(b_t^{AD_f}, b_t^{p_i})$ respectively denote the lateral and longitudinal Euclidean distances between the bounding boxes of the ego vehicle and the $i$-th participant at time $t$. $v.x$ and $v.y$ represent the lateral and longitudinal speeds respectively. $ac.x$ is the lateral acceleration and $ac.y$ is the longitudinal acceleration. The larger $I(fp, ts)$, the higher the fitness score.}


\thx{(2) Objective 2: the motion deviation between $AD_f$ and $AD_o$. The motion deviation between $AD_f$ and $AD_o$ in the same test scenario $ts$, is defined as:}
\thx{
\begin{equation}
L(fp, ts) = \sum_{t=1}^{n} \sqrt{(x_{AD_f}^{t}-x_{AD_o}^{t})^2+(y_{AD_f}^{t}-y_{AD_o}^{t})^2}
\end{equation}}
\thx{where $(x_{AD_f}^{t}, y_{AD_f}^{t})$ represents the position of $AD_f$'s waypoint at time $t$ in the scenario $ts$. The larger $L(fp, ts)$, the higher the fitness score of the test scenario.} 

\noindent\thx{\textbf{Variation Operators.}}
\thx{The variation consists of two operators: crossover and mutation.}

\thx{(1) Crossover. It is applied between two individuals. For each parent individual, \name randomly generates a crossover rate $r_c \in (0,1)$. If $r_c > threshold_c$, the crossover is performed. Specifically, for a single sensor fault, \name applies uniform crossover on chromosomes across the two individuals. For sensor co-fault, it exchanges chromosomes corresponding to the same sensor fault.}



\thx{(2) Mutation. This is applied inside an individual. For each parent individual, \name randomly generates a mutation rate $r_m \in (0,1)$. If $r_m > threshold_m$, gene mutation is performed by modifying one parameter of a sensor. 
Specifically, \name randomly selects one gene from one chromosome and applies Gaussian mutation, introducing adaptive perturbations to efficiently explore the sensor fault model’s parameter space.}


\subsubsection{\thx{Comparing Results}}
\thx{During the execution of test scenarios, \name continuously monitors and records the states of the ego vehicle and participants in real-time, including the waypoint sequences of the ego vehicle and participants. Each waypoint is recorded as a 4-tuple [x-coordinate, y-coordinate, speed, orientation]. \name builds the test oracle by a formal safety specification to compare the execution results based on the recorded data. In test scenario $ts$, the safety violation of the ADS caused by an instance $fp$ of the injected sensor fault $sf$, is identified by $SVF(RS_{fp}^{ts})$: }
\thx{\begin{equation}
    SVF(RS_{fp}^{ts}) = \Phi_f(rs_{AD_f}^{ts}) \wedge \Phi_o(rs_{AD_o}^{ts})
\end{equation}
\begin{equation}
    \Phi_f(rs_{AD_f}^{ts}) = \mathbf{F}_{t \in [0, L]} (Co(rs_{AD_f}^{ts}.t) \vee TS(rs_{AD_f}^{ts}.t) \vee TV(rs_{AD_f}^{ts}.t) \vee TD(rs_{AD_f}^{ts}.L))
\end{equation}
\begin{equation}
    \Phi_o(rs_{AD_o}^{ts}) = \mathbf{G}_{t \in [0, L]} (\neg Co(rs_{AD_o}^{ts}.t) \wedge \neg TS(rs_{AD_o}^{ts}.t) \wedge \neg TV(rs_{AD_o}^{ts}.t) \wedge \neg TD(rs_{AD_o}^{ts}.t))
\end{equation}}
\thx{where $RS_{fp}^{ts}$ records the execution results of $AD_f$ and $AD_o$ in $ts$, represented as $rs_{AD_f}^{ts}$ and $rs_{AD_o}^{ts}$ respectively. $rs_{AD_f}^{ts}.t$ denotes the state of $AD_f$ at time step $t$ in $ts$, including its position, velocity, orientation, and relative distances to participants. $rs_{AD_o}^{ts}.t$ denotes the state of $AD_o$ at $t$ in $ts$. $\Phi_f(rs_{AD_f}^{ts})$ is defined as a temporal logic formula, which asserts whether $AD_f$ violates any safety specification during the execution time of $ts$. Similarly, $\Phi_o(rs_{AD_o}^{ts})$ is to assert whether $AD_o$ violates no safety specification during the execution time of $ts$. $Co, TS, TV, TD$ are the safety specifications. Their descriptions for $AD_f$ are given below, which are similar as those for $AD_o$.}
\thx{\begin{itemize}
    \item $Co(rs_{AD_f}^{ts}.t)$ represents $AD_f$ colliding with any object in $ts$ at time $t$. 
    \item $TS(rs_{AD_f}^{ts}.t)$ represents $AD_f$ blocking or interrupting any participant in $ts$ at time $t$ (comparing to the normal driving of the participant in the scenario where $AD_o$ operates).
    \item $TV(rs_{AD_f}^{ts}.t)$ represents the speed of $AD_f$ exceeding the speed limit of the road or $AD_f$ running the red light at an intersection in $ts$ at time $t$.  
    \item $TD(rs_{AD_f}^{ts}.t)$ represents $AD_f$ failing to arrive at the destination where $AD_o$ arrives at the ending time of $ts$. 
\end{itemize}}
\thx{Note that for $ts$, among the instances $fp$ generated by the GA-based optimization of $sf$, if the number of $fp$ whose $SVF(RS_{fp}^{ts})$ is true, is more than $M$, $sf$ is considered to be capable of inducing safety violations of the ADS in $ts$.}

\section{Evaluation}
To comprehensively evaluate \name, we explore the following research questions:
\begin{itemize}[leftmargin=*]
    \item \textbf{RQ1:} \thx{Can \name discover sensor fault-tolerance issues in ADSs against various sensor faults?}
    \item \thx{\textbf{RQ2:} How effective is \name in sensor fault modeling and injection compared to baselines?}
    \item \textbf{RQ3:} How \thx{effective} is the differential fuzzer-based sensor fault-tolerance testing of \name? 
    \item \textbf{RQ4:} What are the practical impacts of sensor faults injected into ADSs in the physical world?
\end{itemize}

\subsection{Experiment Settings}
\textbf{ADS Under Test.}
We select an industry-grade MSF-based Level-4 ADS, Baidu Apollo \cite{apolloauto} as the test target. Apollo leverages MSF perception to recognize and understand objects in the surrounding environment, primarily integrating data from camera and LiDAR sensors. It has been widely recognized and adopted in the autonomous driving industry with the following evidences. (1) The Apollo community ranks among the top-four leading industrial ADS developers \cite{rank}, while the other three ADSs are not publicly released. (2) Apollo can be readily installed on vehicles for driving on public roads \cite{launch}. \thx{It has been} commercialized for many real-world self-driving services \cite{selfhighway, baidutaxi}. 

\textbf{Simulation Environment.}
We conduct the simulation experiments on Ubuntu 20.04 with 500 GB memory, an Intel Core i9 CPU, and an NVIDIA GTX3090 TI. We adopt SORA-SVL \cite{sora-svl}, an end-to-end AV simulation platform that supports connection with Apollo.

\thx{\textbf{Parameter Settings.} We consider and set the following parameters in \name: (1) $num$: this is the number of generated test scenarios for each injected sensor fault. We set it as 100 for the balance of scenario coverage and time cost. (2) $threshold_m$ and $threshold_c$: these are the thresholds for mutation and crossover, respectively. We test different values of these two parameters recommended by existing genetic algorithms \cite{haupt2000optimum, tian2022mosat}, and choose 0.3 for $threshold_m$ and 0.4 for $threshold_{c}$. (3) $k$ is the number of selected excellent individuals in each generation, and $M$ is the threshold for determining a sensor fault capable of causing ADS's safety violations. To balance the search effects and evolving costs, we set $k$ as 4 and $M$ as 5.}

\subsection{Experiment Design}
\textbf{To answer RQ1}, we apply \name to test Apollo's tolerance of different sorts of sensor faults. For each sensor fault and co-fault, we generate 100 test scenarios, encompassing various driving situations with different types of roads, participants, and weather conditions. Road types include highways, urban streets, and intersections. Participants consist of behaviors of NPC vehicles (including following lanes, changing lanes, crossing, turning around, overtaking, and parking) and pedestrians (including walking along, walking across, and standing). 



\thx{\textbf{To answer RQ2}, considering that there are no available approaches that could test the system-level performance of MSF-based ADSs with injected sensor faults, we select an MSF robustness and reliability testing benchmark \cite{gaobenchmark}, as the baseline for comparison. This benchmark summarizes and implements a range of real-world corruption patterns on MSF perception modules, and tests their impacts on the MSF results in representative perception tasks (e.g., object detection, object tracking, and depth completion). They conclude 14 corruption patterns, which simulate corrupted data of scenarios and input into the perception module to obtain the output result. 12 of these 14 corruption patterns exhibit similar characteristics to a part of the sensor faults implemented in \name, e.g., \textit{rain in environment} in the benchmark corresponds to \textit{rain on lens and LiDAR} in \name. We refer to those overlapping patterns as \textit{common patterns}. }
\thx{We implement the baseline by integrating the common patterns in the benchmark into Apollo, and test its safety with the input of corrupted sensor data. For fairness, we test Apollo with each common pattern in the same test scenarios as \name, and explore it using the differential fuzzer of \name.}


\textbf{To answer RQ3}, we conduct the ablation experiment that compares \name with \nameb, which replaces the differential fuzzer with a random sampling of sensor fault parameters. For the sensor fault $sf$ in the test scenario $ts$, we use the random-based baseline to generate the same number of instances of $sf$ as those generated by \name in $ts$, and test the fault tolerance of Apollo.

Note that for RQ1, RQ2, and RQ3, \thx{to account for the randomness of the differential fuzzer in \name, each experiment is repeated ten times. Meanwhile, across the ten runs of experiments, for RQ1, we vary the parameters' values of the 100 test scenarios considering the randomness of the parameter sampling of test scenarios. For RQ2 and RQ3, we use the same scenarios in RQ1 to test the baselines for fair comparisons.}  

\begin{figure}[h]
  \centering
  \includegraphics[scale=0.2]{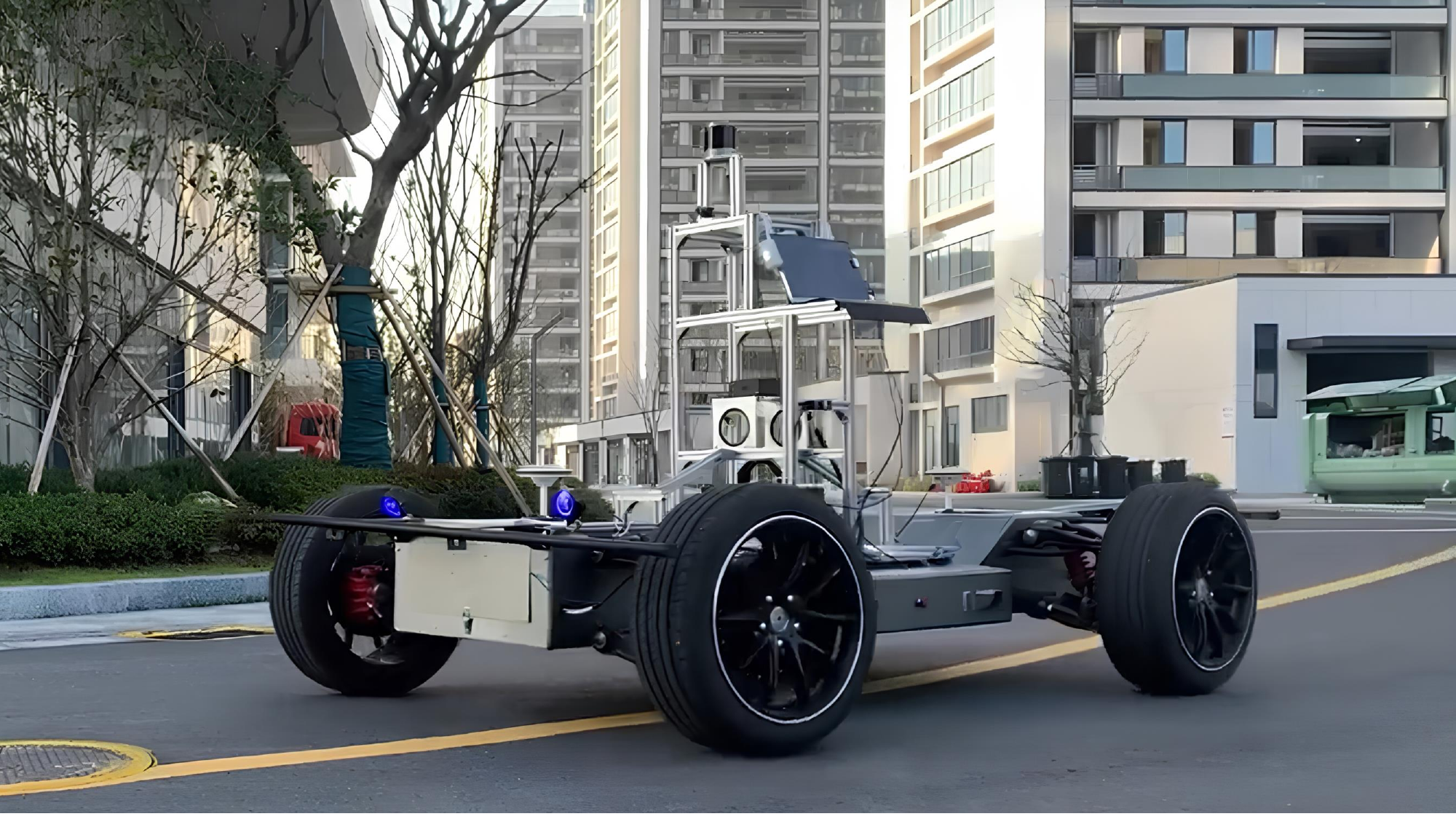}
  \caption{The autonomous vehicle equipped with Apollo 6.0 Edu for our physical experiment.}
  \label{vehicle}
  \end{figure}

\textbf{To answer RQ4}, we conduct the experiments on an actual AV in real-world roads. As illustrated in Figure~\ref{vehicle}, our AV is equipped with a 32-line LiDAR, 1920*1080p HD camera, Huace GI-410 INS, and a Nuvo-8111 industrial PC with an Intel Core i9-9900K CPU, NVIDIA RTX 3060 GPU, 32GB RAM, and 1TB SSD, integrated with Pix Hooke Chassis and Apollo 6.0 Edu Platform.

\subsection{RQ1: Effectiveness Experiment}
\begin{figure}[h]
    \centering
    \begin{subfigure}[b]{0.45\linewidth}
        \centering
        \includegraphics[width=\linewidth]{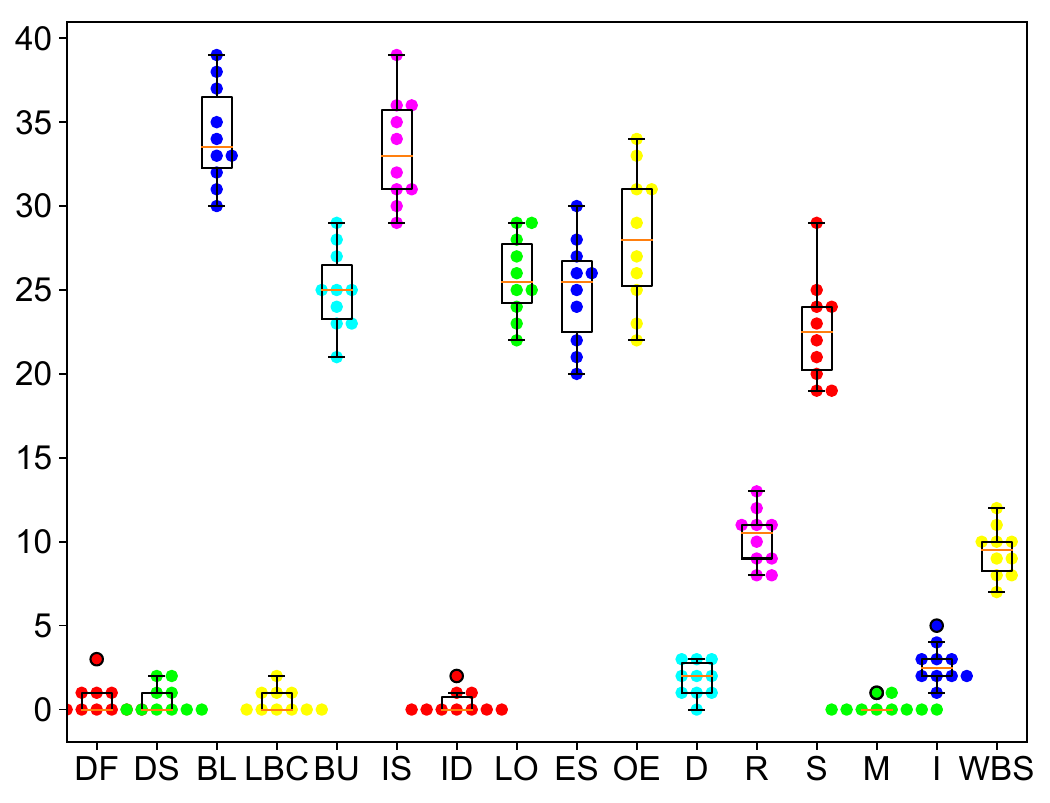}
        \caption{Number of SVs caused by camera faults}
        \label{sub1}
    \end{subfigure}
    \begin{subfigure}[b]{0.45\linewidth}
        \centering
        \includegraphics[width=\linewidth]{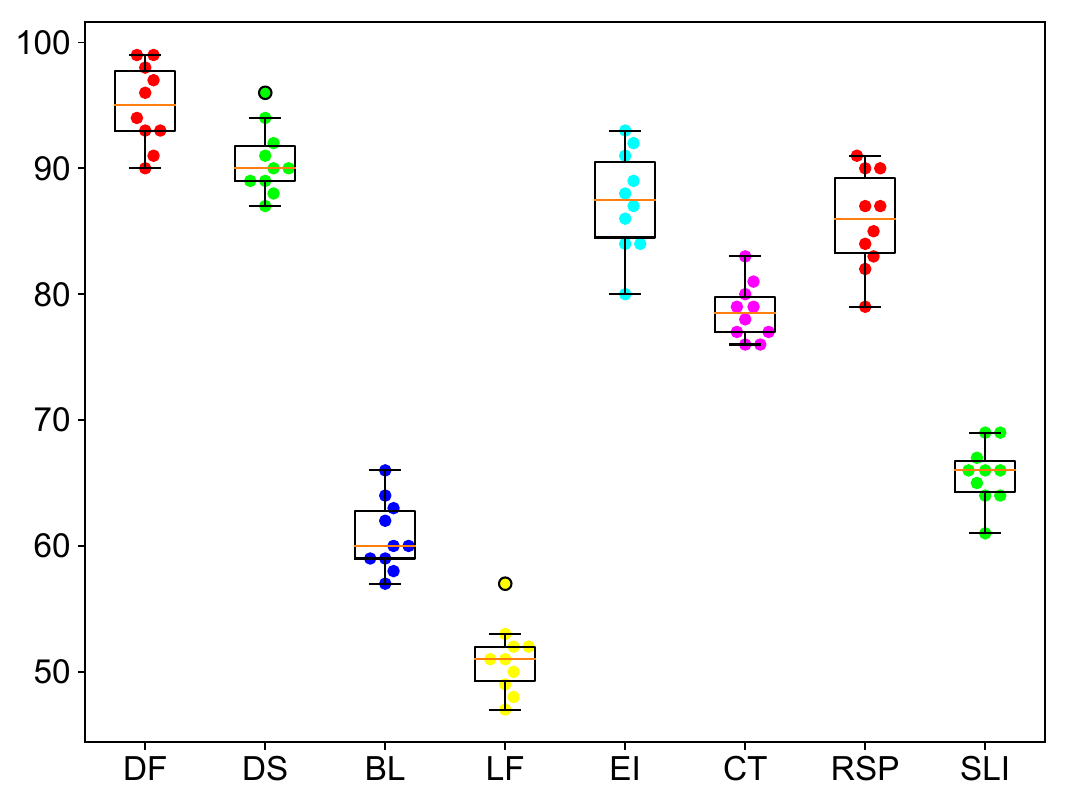}
        \caption{Number of SVs caused by LiDAR faults}
        \label{sub2}
    \end{subfigure}
    \caption{The number of safety violations and variance under camera and LiDAR fault injection}
    \label{comparison_box}
\end{figure}

Our effectiveness evaluation results are shown in Tables~\ref{camera-fault} and \ref{lidar-fault}, where SV is short for "safety violation". We also show the average numbers of SVs and variance for each fault in Figure~\ref{comparison_box} (where the sequences of camera and LiDAR faults in the $x$-axis correspond to the ones in Tables~\ref{camera} and \ref{lidar}). We have two high-level observations. First, MSF exhibits different degrees of fault tolerance against different types of faults. Second, for each run of the experiments, the results remain consistent, implying that the camera and LiDAR faults have predictable and deterministic impacts on the ADS's behaviors. 
Some experiment videos are available at \url{https://zenodo.org/uploads/14015455}. Below we present more in-depth analysis and findings about MSF's fault tolerance. 

\begin{table}[h]\footnotesize
\caption{The \thx{average} numbers of safety violations caused by camera faults}
\begin{tabular}{|c|c|c|c|c|c|c|c|c|}
\hline
\textbf{\begin{tabular}[c]{@{}c@{}}Fault\\ Type\end{tabular}}    & \begin{tabular}[c]{@{}c@{}}Deflec-\\ tion\end{tabular}     & \begin{tabular}[c]{@{}c@{}}Displace-\\ ment\end{tabular} & \begin{tabular}[c]{@{}c@{}}Broken \\ Lens\end{tabular} & \begin{tabular}[c]{@{}c@{}}Lens Bright-\\ ness Change\end{tabular} & Blur & \begin{tabular}[c]{@{}c@{}}Internal\\ Scatter\end{tabular} & \begin{tabular}[c]{@{}c@{}}Internal\\ Dirt\end{tabular} & \begin{tabular}[c]{@{}c@{}}Lens Occl-\\ usion\end{tabular}       \\ \hline
\textbf{\begin{tabular}[c]{@{}c@{}}Number\\ of SVs\end{tabular}} & 0.6                                                        & 0.6                                                      & 34.2                                                  & 0.5                                                               & 25   & 33.3                                                       & 0.4                                                     & 25.8                                                            \\ \hline
\textbf{\begin{tabular}[c]{@{}c@{}}Fault\\ Type\end{tabular}}    & \begin{tabular}[c]{@{}c@{}}External\\ Scatter\end{tabular} & \begin{tabular}[c]{@{}c@{}}Overexpo-\\ sure\end{tabular} & Dust                                                  & Rain                                                              & Snow & Mist                                                       & Ice                                                     & \begin{tabular}[c]{@{}c@{}}White Bal-\\ance Shift\end{tabular} \\ \hline
\textbf{\begin{tabular}[c]{@{}c@{}}Number\\ of SVs\end{tabular}} & 24.9                                                       & 28.1                                                     & 1.8                                                   & 10.2                                                              & 22.6 & 0.2                                                        & 2.7                                                     & 9.4                                                             \\ \hline
\end{tabular}
\label{camera-fault}
\end{table}

\subsubsection{Camera Fault.}
The ADS exhibits significant different behaviors under various types of camera faults. 
  Among the sixteen camera faults, we observe that only half of them can cause a large number of safety violations in Apollo, and their impacts on Apollo's behaviors vary greatly.

First, from Table~\ref{camera-fault} we observe that some camera faults, including deflection, displacement, lens brightness change, internal dirt, dust, mist, and ice, do not frequently lead to safety violations of Apollo, which suggests that the MSF perception-based ADS demonstrates a notable degree of resilience to such slight distortions. Specifically, the perception pipeline can tolerate slight distortions in image quality or slight obstructions in the camera lens well, such as slight shifts in camera alignment (deflection or displacement) or gradual brightness changes, because the complementary LiDAR input can mitigate visual defects. Additionally, small particles or slight occlusions (such as dust or fog) on the camera lens may not block a large part of the field of view, allowing the ADS to maintain a consistent scenario interpretation.


\begin{center}
\begin{tcolorbox}[colback=gray!15,
                  colframe=black,
                  width=14cm,
                  arc=1mm, auto outer arc,
                  boxrule=0.5pt,size=title,opacityfill=0.1
                 ]
\textbf{Finding 1:} 
The MSF perception-based ADS has strong fault tolerance on unstructured visual disturbances to the camera.
\end{tcolorbox}
\end{center}

Finding 1 indicates that the MSF perception-based ADS is inherently robust to low-level visual noise or occlusions, ensuring that minor disturbances do not trigger erroneous system behaviors. This resilience is critical in real-world applications, where \textit{minor visual obstructions are inevitable due to varying environmental conditions.} Consequently, the ADS’s capacity to manage these subtle distortions suggests that multi-sensor fusion frameworks enhance fault tolerance and contribute to safer operational performance by reducing the impact of minor camera faults.

Second, some camera faults, including broken lens, blur, internal and external scatter, lens occlusion, and overexposure, can distort the structural integrity of image. They introduce significant distortions to the image data, affecting critical visual features (e.g., object edges, textures, spatial clarity). From Table~\ref{camera-fault}, we observe that these faults can lead to more safety violations in Apollo. Such safety violations occur when the motion of objects near Apollo changes. These faults can lead to MSF's misdetection and misclassification of objects in scenarios. Furthermore, they disrupt the ADS’s ability to perform precise spatial localization and safe navigation. For example, a broken lens or severe blurring can obscure the boundaries of objects, making it difficult for the ADS to discern the presence or position of pedestrians, vehicles, or road obstacles. Overexposure and lens occlusion faults exacerbate this issue by introducing areas of high brightness or visual blockage, leading to a limited field of view that prevents the ADS from obtaining a complete and reliable representation of its surroundings. As a result, these structural distortions not only degrade image quality but also lead to frequent safety violations as the ADS makes incorrect or delayed decisions.

\begin{center}
\begin{tcolorbox}[colback=gray!15,
                  colframe=black,
                  width=14cm,
                  arc=1mm, auto outer arc,
                  boxrule=0.5pt,size=title,opacityfill=0.1
                 ]
\textbf{Finding 2:} Camera faults that significantly compromise the structural integrity of the visual data can easily cause misbehaviors of ADSs in response to motion changes of nearby objects.
\end{tcolorbox}
\end{center}


Third, the impacts of passive camera faults (e.g., snow, ice, raindrops, mist, or dust on the lens) on ADS's behaviors vary greatly. Specifically, heavy snow and ice on the lens tend to accumulate in larger volumes, creating substantial occlusion on the lens and severely obstructing light transmission, which can change the structural aspects of the visual data. Raindrops, although typically smaller in volume than snow and ice, possess unique optical properties that cause the light to refract and scatter as it passes through or around the droplets. This scattering effect can distort object shapes and positions, leading to erroneous behaviors of the ADS. In contrast, mist and dust on the lens form a thin, diffuse layer that moderately reduces the light transmittance but with a relatively lower scattering rate, often resulting in a softened image that preserves object outlines and only reduces the contrast and detail. So MSF-based ADSs can resolve them well.

\begin{center}
\begin{tcolorbox}[colback=gray!15,
                  colframe=black,
                  width=14cm,
                  arc=1mm, auto outer arc,
                  boxrule=0.5pt,size=title,opacityfill=0.1
                 ]
\textbf{Finding 3:} The passive faults that cause contamination to the lens have significant different impacts on ADS's behaviors, which are related to the inherent properties of the pollutants (e.g., volume, transmittance, and scattering rate).
\end{tcolorbox}
\end{center}

Fourth, the white balance shift can only cause safety violations of Apollo during driving in two traffic situations. When this fault causes the camera background color to be light red, the MSF perception identifies the green light as a red light, causing the ADS to stop at the intersection and disrupt the traffic flow. When the white balance shift causes the camera background color to be light yellow, the MSF-based perception recognizes the yellow light as a green light, causing the ADS's decision of running through even if the ego vehicle \thx{does} not exceed the stopping line, which violates the traffic regulation and increases the risk of collisions at intersections. 

\begin{center}
\begin{tcolorbox}[colback=gray!15,
                  colframe=black,
                  width=14cm,
                  arc=1mm, auto outer arc,
                  boxrule=0.5pt,size=title,opacityfill=0.1,size=title,opacityfill=0.1
                 ]
\textbf{Finding 4:} White balance shift can affect ADS's accuracy in identifying traffic lights under specific circumstances.
\end{tcolorbox}
\end{center} 

\begin{table}[h]\small
\caption{The \thx{average} numbers of safety violations caused by LiDAR faults}
\begin{tabular}{|c|c|c|c|c|}
\hline
\textbf{\begin{tabular}[c]{@{}c@{}}Fault\\ Type\end{tabular}} & Deflection                                                             & Displacement & Beam Loss                                                         & Line Fault                                                          \\ \hline
\textbf{Number of SVs}                                        & 95                                                                     & 90.6         & 60.8                                                              & 51                                                                  \\ \hline
\textbf{\begin{tabular}[c]{@{}c@{}}Fault\\ Type\end{tabular}} & \begin{tabular}[c]{@{}c@{}}Electromagnetic\\ Interference\end{tabular} & Crosstalk    & \begin{tabular}[c]{@{}c@{}}Rain and Snow\\ Pollution\end{tabular} & \begin{tabular}[c]{@{}c@{}}Strong Light\\ Interference\end{tabular} \\ \hline
\textbf{Number of SVs}                                        & 87.4                                                                   & 78.6         & 85.8                                                              & 65.7                                                                \\ \hline
\end{tabular}
\label{lidar-fault}
\end{table}


\subsubsection{LiDAR faults.}
The LiDAR faults have much stronger impacts than camera faults on MSF-based ADSs. As shown in Table~\ref{lidar-fault}, nearly all types of LiDAR faults are capable of inducing erroneous behaviors of ADSs.
The point clouds captured by LiDAR provide detailed 3D spatial context, and are directly used by the ADS for obstacle recognition and avoidance. Camera faults, while potentially affecting object recognition in certain visual conditions, do not have \thx{a critical impact} on real-time obstacle avoidance.

\begin{center}
\begin{tcolorbox}[colback=gray!15,
                  colframe=black,
                  width=14cm,
                  arc=1mm, auto outer arc,
                  boxrule=0.5pt, size=title,opacityfill=0.1
                 ]
\textbf{Finding 5:} The MSF perception-based ADS has low fault tolerance of LiDAR faults.
\end{tcolorbox}
\end{center} 

The most significant distinction between camera faults and LiDAR faults lies in how deflection and displacement influence ADS behaviors and create system-level consequences. Specifically, while deflection and displacement in the camera rarely lead to safety violations, similar faults in the LiDAR are among the highest in inducing safety-critical failures. We analyze \thx{the} main reason for such difference: when a deflection or displacement fault occurs in LiDAR, it skews these distance measurements and point cloud data. As a result, the ADS experiences a fundamental misinterpretation of object positions and distances. Such inaccuracies have a direct and high-impact consequence on the decision-making of the ADS. The camera only \thx{contributes} to the 2D spatial understanding by supplementing object classification and contextual information. This makes the camera's spatial misalignments less impactful, and MSF-based perception can correct this fault by the MSF algorithm and LiDAR data.
Therefore, the deflection and displacement faults in camera have a low likelihood of propagating severe errors into the decision-making process. Compared to the camera, when LiDAR data is corrupted due to deflection or displacement, it introduces inconsistencies in the sensor fusion results, which cannot be resolved well by the MSF algorithm.

\begin{center}
\begin{tcolorbox}[colback=gray!15,
                  colframe=black,
                  width=14cm,
                  arc=1mm, auto outer arc,
                  boxrule=0.5pt,size=title,opacityfill=0.1
                 ]
\textbf{Finding 6:} The MSF perception-based ADS is more sensitive to LiDAR deflection and displacement than to camera misalignment.
\end{tcolorbox}
\end{center} 


Besides, the beam loss and line fault are the conditions of LiDAR aging and degradation, where single laser beams weaken or malfunction. These faults result in reduced partial density in the point cloud, compromising the system’s ability to detect smaller or distant objects reliably. Compared with other faults of LiDAR, Beam loss and line fault causes fewer safety violations. This suggests that the MSF-based ADS is tolerant against aging and degradation to some extent. Such fault tolerance may remain until the long-term aging and degradation reach a critical threshold. As aging and degradation faults of LiDAR develop over extended periods, the MSF perception-based ADS can adapt incrementally as the sensor’s performance gradually shifts. This progressive nature makes it possible for the ADS to identify and compensate for these slow changes. \thx{Specifically, the} design of MSF perception-based ADS typically incorporates sensor redundancy. The ADS can adjust MSF perception to recalibrate its confidence weights for LiDAR data, relying more on input from other sensors when a decrease in LiDAR reliability is detected.

\begin{center}
\begin{tcolorbox}[colback=gray!15,
                  colframe=black,
                  width=14cm,
                  arc=1mm, auto outer arc,
                  boxrule=0.5pt,size=title,opacityfill=0.1
                 ]
\textbf{Finding 7:} The MSF perception-based ADS exhibits a better fault tolerance for LiDAR aging and degradation than for external interferences.
\end{tcolorbox}
\end{center}

\subsubsection{Co-faults}
\begin{figure}[h]
  \centering
  \includegraphics[scale=0.8]{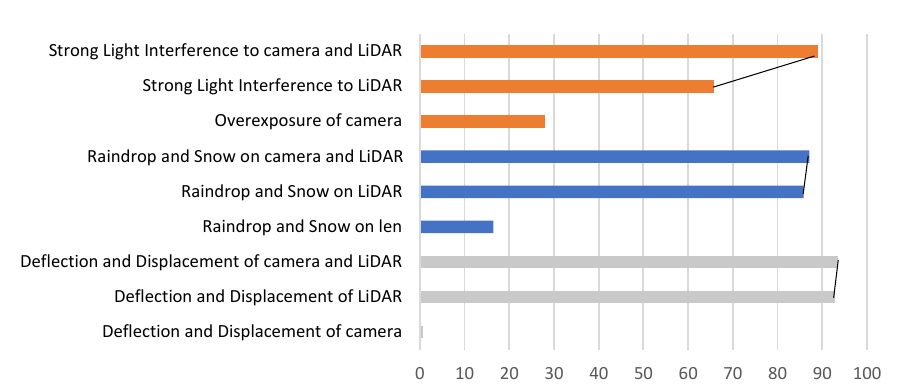}
  \caption{The average numbers of safety violations of Apollo caused by different co-faults.}
  \label{combination}
  \end{figure}

Next we consider the combination of multiple sensor faults triggered by the same condition. This includes: deflection and displacement (bumpy roads), raindrops and snow on sensors (sleety weather), overexposure and strong light interference (under high beam). The safety violations of Apollo caused by these combinations are shown in Figure~\ref{combination}.

The MSF-based ADS leverages complementary data from both the camera and LiDAR, effectively mitigating the impact of missing different partial information from one sensor alone. In situations where partial data loss occurs in either the camera or LiDAR input, the ADS can compensate for it by relying on the other sensor to retrieve the data for the missing part. This complementary data enables the localization, prediction, and decision-making modules of the ADS to generate accurate and reliable operational commands, thus sustaining the vehicle’s safe navigation and decision-making capabilities.

However, under strong light conditions, both the camera and LiDAR sensors experience concurrent data degradation simultaneously, which overexposes camera inputs and saturates LiDAR sensors.
This lack of alternative input due to identical data loss from the camera and LiDAR may cause the ADS's localization and prediction modules to fail to detect obstacles and misinterpret distances, or generate inaccurate predictions of nearby objects' movements. Consequently, the decision-making module, which relies on accurate perception data, may issue incorrect operational commands, which critically undermine the correctness and safety of ADS’s behaviors. 

\begin{center}
\begin{tcolorbox}[colback=gray!15,
                  colframe=black,
                  width=14cm,
                  arc=1mm, auto outer arc,
                  boxrule=0.5pt,size=title,opacityfill=0.1
                 ]
\textbf{Finding 8:} Strong light interference that simultaneously disrupts both camera and LiDAR significantly increases the frequency of safety violations of MSF perception-based ADSs.
\end{tcolorbox}
\end{center} 

\subsection{RQ2: Comparison Experiment}
\thx{Table~\ref{baseline} shows the sensor faults in \name that correspond to the corruption patterns in the baseline benchmark, as well as their number of violations. Note that there exist some sensor faults that contain more than one corruption pattern (e.g., both \textit{Motion Blur} and \textit{Defocus Blur} in the benchmark are included in \textit{Blur} of \name ). For these sensor faults, we accumulate the impacts of the corresponding corruption patterns as the baseline's input data of test scenarios.}

\begin{table}[h]\small
\caption{Comparison results of \name and the baseline benchmark}
\thx{
\begin{tabular}{|c|c|cccc|c|cccc|}
\hline
\multirow{2}{*}{\textbf{\begin{tabular}[c]{@{}c@{}}Sensor\\ Type\end{tabular}}} & \multirow{2}{*}{\textbf{\begin{tabular}[c]{@{}c@{}}Corruption in\\ Benchmark\end{tabular}}} & \multicolumn{4}{c|}{\textbf{\begin{tabular}[c]{@{}c@{}}Number\\ of SVs\end{tabular}}}                       & \multirow{2}{*}{\textbf{\begin{tabular}[c]{@{}c@{}}Sensor Faults\\ in \name\end{tabular}}} & \multicolumn{4}{c|}{\textbf{\begin{tabular}[c]{@{}c@{}}Number\\ of SVs\end{tabular}}}                       \\ \cline{3-6} \cline{8-11} 
                                                                                &                                                                                             & \multicolumn{1}{l|}{avg}  & \multicolumn{1}{l|}{min} & \multicolumn{1}{c|}{md}   & \multicolumn{1}{l|}{max} &                                                                                           & \multicolumn{1}{l|}{avg}  & \multicolumn{1}{l|}{min} & \multicolumn{1}{c|}{md}   & \multicolumn{1}{l|}{max} \\ \hline
\multirow{4}{*}{Camera}                                                         & \begin{tabular}[c]{@{}c@{}}Brightness\\ Increasing\end{tabular}                             & \multicolumn{1}{c|}{9.6}  & \multicolumn{1}{c|}{6}   & \multicolumn{1}{c|}{9.5}  & 12                       & Overexposure                                                                              & \multicolumn{1}{c|}{28.1} & \multicolumn{1}{c|}{22}  & \multicolumn{1}{c|}{28}   & 34                       \\ \cline{2-11} 
                                                                                & \begin{tabular}[c]{@{}c@{}}Darkness\end{tabular}                 & \multicolumn{1}{c|}{0.4}  & \multicolumn{1}{c|}{0}   & \multicolumn{1}{c|}{1}    & 1                        & \begin{tabular}[c]{@{}c@{}}Lens Brightness\\ Change\end{tabular}                          & \multicolumn{1}{c|}{0.5}  & \multicolumn{1}{c|}{0}   & \multicolumn{1}{c|}{0}    & 1                        \\ \cline{2-11} 
                                                                                & Image Noise                                                                                 & \multicolumn{1}{c|}{18.6} & \multicolumn{1}{c|}{11}  & \multicolumn{1}{c|}{18}   & 29                       & Internal Scatter                                                                          & \multicolumn{1}{c|}{33.3} & \multicolumn{1}{c|}{29}  & \multicolumn{1}{c|}{33}   & 39                       \\ \cline{2-11} 
                                                                                & \begin{tabular}[c]{@{}c@{}}Motion\&De-\\focus Blur\end{tabular}                              & \multicolumn{1}{c|}{18.5} & \multicolumn{1}{c|}{9}   & \multicolumn{1}{c|}{18}   & 24                       & Blur                                                                                      & \multicolumn{1}{c|}{25}   & \multicolumn{1}{c|}{21}  & \multicolumn{1}{c|}{25}   & 29                       \\ \hline
\multirow{3}{*}{LiDAR}                                                          & \begin{tabular}[c]{@{}c@{}}Point Cloud\\ Gaussian Noise\end{tabular}                        & \multicolumn{1}{c|}{41.1} & \multicolumn{1}{c|}{30}  & \multicolumn{1}{c|}{43.5} & 47                       & Line Fault                                                                                & \multicolumn{1}{c|}{51}   & \multicolumn{1}{c|}{47}  & \multicolumn{1}{c|}{51}   & 57                       \\ \cline{2-11} 
                                                                                & \begin{tabular}[c]{@{}c@{}}Point Cloud\\ Impulse Noise\end{tabular}                         & \multicolumn{1}{c|}{60.5} & \multicolumn{1}{c|}{48}  & \multicolumn{1}{c|}{63}   & 69                       & \begin{tabular}[c]{@{}c@{}}Electromagnetic\\ Interference\end{tabular}                    & \multicolumn{1}{c|}{87.4} & \multicolumn{1}{c|}{80}  & \multicolumn{1}{c|}{87.5} & 93                       \\ \cline{2-11} 
                                                                                & Signal Loss                                                                                 & \multicolumn{1}{c|}{50.6} & \multicolumn{1}{c|}{40}  & \multicolumn{1}{c|}{51}   & 59                       & Beam Loss                                                                                 & \multicolumn{1}{c|}{60.8} & \multicolumn{1}{c|}{57}  & \multicolumn{1}{c|}{60}   & 66                       \\ \hline
\multirow{3}{*}{\begin{tabular}[c]{@{}c@{}}Camera+\\ LiDAR\end{tabular}}        & Fog                                                                                         & \multicolumn{1}{c|}{0.3}  & \multicolumn{1}{c|}{0}   & \multicolumn{1}{c|}{0}    & 1                        & Mist                                                                                      & \multicolumn{1}{c|}{0.2}  & \multicolumn{1}{c|}{0}   & \multicolumn{1}{c|}{0}    & 1                        \\ \cline{2-11} 
                                                                                & Rain                                                                                        & \multicolumn{1}{c|}{25.8} & \multicolumn{1}{c|}{15}  & \multicolumn{1}{c|}{25.5} & 35                       & \begin{tabular}[c]{@{}c@{}}Rain on lens\\ \&LiDAR \end{tabular}                                                                      & \multicolumn{1}{c|}{38.7} & \multicolumn{1}{c|}{36}  & \multicolumn{1}{c|}{38.5} & 41                       \\ \cline{2-11} 
                                                                                & \begin{tabular}[c]{@{}c@{}}Spatial\\ Misalignment\end{tabular}                              & \multicolumn{1}{c|}{52.5} & \multicolumn{1}{c|}{45}  & \multicolumn{1}{c|}{52}   & 59                       & \begin{tabular}[c]{@{}c@{}}Deflection and\\ Displacement\end{tabular}                     & \multicolumn{1}{c|}{95.9} & \multicolumn{1}{c|}{93}  & \multicolumn{1}{c|}{95.5} & 99                       \\ \hline
\end{tabular}
}
\label{baseline}
\end{table}

\thx{We observe \name demonstrates more effectiveness in safety violation identification than the baseline benchmark. Specifically, for the common patterns that have significant impacts on Apollo, \name can discover a larger number of safety violations than the benchmark. For the common patterns that rarely lead to safety violations of Apollo, such as darkness (lens brightness change) and fog (mist), both \name and the benchmark discover almost the same number of safety violations. The results demonstrate that the sensor fault injection of \name is significantly more effective in discovering safety violations than the benchmark.} 

\subsection{RQ3: Ablation Experiment}
\begin{figure}[h]
  \centering
  \includegraphics[scale=0.44]{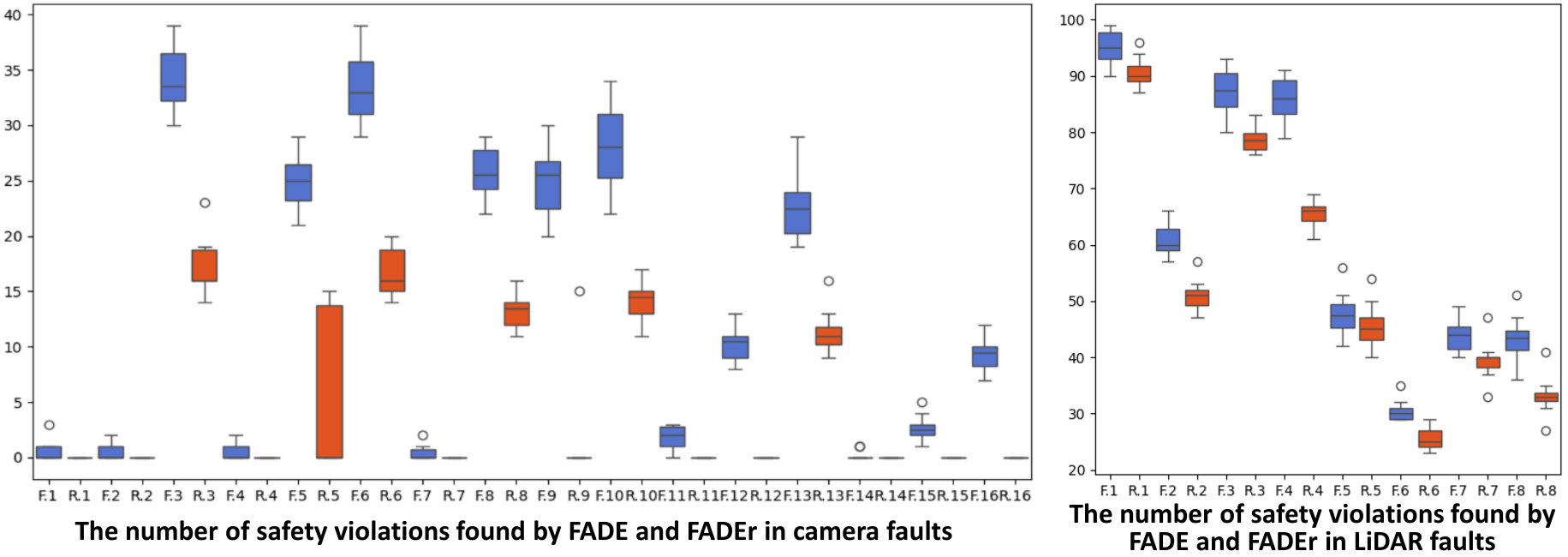}
  \caption{Comparisons of \name and \nameb.}
  \label{comparison_random}
  \end{figure}

\thx{Figure~\ref{comparison_random} shows the comparison results of \name (blue bars in $F.x$) and \nameb (orange bars in $R.x$), where the sequences of camera and LiDAR faults in the $x$-axis correspond to the ones in Tables~\ref{camera} and \ref{lidar}. Overall, compared to \nameb, under the same test scenarios, \name can discover more safety violations of Apollo caused by injected sensor faults.}

\thx{Specifically, for each camera fault, in each run, \name consistently identifies more types of critical faults and safety violations of Apollo. For LiDAR faults, \nameb identifies all types of LiDAR faults that can induce safety violations of Apollo. However, for each fault, the number of Apollo's safety violations detected by \nameb is significantly lower than that by \name. We conclude that \name exhibits superior performance in discovering critical camera faults, as well as safety violations under both camera and LiDAR faults. The results demonstrate that the differential fuzzer of \name is more effective and efficient in exploring the space of sensor faults, and more comprehensive in testing the fault-tolerance of MSF-based ADSs.}


\subsection{RQ4: Physical Experiment}
\thx{To ensure adherence to safety concerning of AV integrity, equipment protection, and safety of the test site and participants, we select the following sensor faults based on the results of RQ1, which can significantly cause safety violations of Apollo in simulation experiments: camera faults of lens occlusion and external scatter, LiDAR faults of deflection and displacement, camera overexposure and LiDAR interference due to strong light. For each selected sensor fault, we randomly choose one of Apollo’s safety violations identified by our approach from each of the 10 runs in RQ1. To ensure diversity, the selected 10 cases have distinct parameter values in the sensor fault model and test scenarios, reducing the risk of occasionality in the physical evaluation. This selection process is repeated five times to account for randomness.}

\thx{For each selected sensor fault, we validate whether the found safety violations of Apollo caused by the injected sensor fault will occur in the physical world. The sensor faults are applied to the actual AV in the following ways, to replicate the simulated counterpart as closely as possible:}
\begin{itemize}
    \item \thx{\textbf{Lens Occlusion:} we use paper sheets to cover specific portions of the lens, ensuring that the occluded area and position match the safety-violation scenarios caused by lens occlusion.}
    \item \thx{\textbf{External Scatters:} according to the safety-violation scenarios caused by external scatters, we disperse colored confetti at predefined densities and locations on the lens of the camera.}
    \item \thx{\textbf{LiDAR Deflection:} we manually adjust LiDAR's orientation angle to introduce specific deflection in safety-violation scenarios caused by LiDAR deflection.}
    \item \thx{\textbf{LiDAR Displacement:} we modify LiDAR's mounting position according to safety-violation scenarios caused by LiDAR displacement, while ensuring that LiDAR is securely attached to the ego vehicle.}
    \item \thx{\textbf{Strong Light Interference:} we use a high-intensity flashlight to project specific intensity of illumination from a predefined angle above the vehicle’s front, replicating the impacts of strong light interference on camera and LiDAR.}
\end{itemize}

The results of physical experiments are shown as Table~\ref{physical}.
The AV failure rate (AFR) for sensor fault $sf$ is represented as $AFR_{sf}$, calculated as $AFR_{sf} = \frac{N_{SV}}{N_{sf}}$, where $N_{SV}$ is the number of safety-violation scenarios induced by $sf$ successfully reproduced in the physical experiment, and $N_{sf}$ is the number of selected safety-violation scenarios induced by $sf$ in simulation testing.

\begin{table}[h]\small
\centering
\caption{The evaluation results of physical experiments}
\renewcommand\arraystretch{1.2}
\begin{tabular}{|c|cc|cc|cc|cc|cc|}
\hline
\textbf{\begin{tabular}[c]{@{}c@{}}Sensor\\ Fault\end{tabular}}                       & \multicolumn{2}{c|}{\begin{tabular}[c]{@{}c@{}}Lens Occlu-\\ sion\end{tabular}} & \multicolumn{2}{c|}{\begin{tabular}[c]{@{}c@{}}External\\ Scatter\end{tabular}} & \multicolumn{2}{c|}{\begin{tabular}[c]{@{}c@{}}LiDAR\\ Deflection\end{tabular}} & \multicolumn{2}{c|}{\begin{tabular}[c]{@{}c@{}}LiDAR\\ Displacement\end{tabular}} & \multicolumn{2}{c|}{\begin{tabular}[c]{@{}c@{}}Strong Light to\\ Camera and LiDAR\end{tabular}} \\ \hline
\multirow{3}{*}{\textbf{\begin{tabular}[c]{@{}c@{}}AV\\ Failure\\ Rate\end{tabular}}} & \multicolumn{1}{c|}{min}                         & 50\%                        & \multicolumn{1}{c|}{min}                         & 50\%                         & \multicolumn{1}{c|}{min}                         & 70\%                         & \multicolumn{1}{c|}{min}                          & 70\%                          & \multicolumn{1}{c|}{min}                                 & 80\%                                 \\ \cline{2-11} 
                                                                                      & \multicolumn{1}{c|}{avg}                         & 60\%                        & \multicolumn{1}{c|}{avg}                         & 62\%                         & \multicolumn{1}{c|}{avg}                         & 82\%                         & \multicolumn{1}{c|}{avg}                          & 80\%                          & \multicolumn{1}{c|}{avg}                                 & 92\%                                 \\ \cline{2-11} 
                                                                                      & \multicolumn{1}{c|}{max}                         & 70\%                        & \multicolumn{1}{c|}{max}                         & 70\%                         & \multicolumn{1}{c|}{max}                         & 90\%                         & \multicolumn{1}{c|}{max}                          & 90\%                          & \multicolumn{1}{c|}{max}                                 & 100\%                                \\ \hline
\end{tabular}
\label{physical}
\end{table}

In the five runs of physical experiments, LiDAR faults cause higher AFRs than camera faults: the former leads to an average AFR of more than 80\%, and the AFR of the later can also reach over 60\% and the minimum is 50\%. The AFR of overexposure and strong light interference ranks the highest. These physical experiment results demonstrate that the safety violations of the ADS caused by sensor faults discovered by \name have strong practical significance. The effects of sensor faults are reliable and deterministic in real-world environments, mirroring their behaviors in simulations. 

\section{Threats to Validity}


\noindent\textbf{Selective validation of physical experiments.} One primary threat is that not all sensor faults are injected into the physical AV to validate their impacts on the ADS in the real world. Due to sensor component costs and the safety of vehicles and pedestrians in physical experiments, we selectively validate those sensor faults where the sensor components will not be damaged. For the safety of vehicles and pedestrians involved in physical experiments, we choose cardboard boxes to replace participants in simulation scenarios. While this threat exists, we conduct an in-depth analysis of the underlying causes of our findings, which can provide convincing recommendations for developers, testers, and safety researchers in the field of ADSs.

\noindent\thx{\textbf{Parameter ranges of sensor fault models.} Another potential threat is that the difficulty of identifying safety violations caused by the injected sensor faults could depend on the parameter ranges of the fault model. The GA-based differential fuzzer of \name inherently explores the parameter spaces of each sensor fault model, which can adapt to varying ranges of them. Consequently, regardless of the size of the parameter ranges, \name guarantees thorough exploration of sensor fault models. We plan to conduct a detailed analysis of parameter range variations for sensor fault models, exploring the robustness of ADS to different parameter ranges of sensor fault models.}

\section{Conclusion}
\thx{This paper proposes \name}, the first approach to test the fault tolerance of MSF-based ADSs against different types of sensor faults. \name designs sensor fault models for injecting camera and LiDAR faults into the MSF-based ADS, and implements a GA-guided differential fuzzer to explore the parameter spaces of sensor fault models. We evaluate \name on an industrial MSF-based ADS. The evaluation results demonstrate that \name can effectively and efficiently discover Apollo's safety violations caused by the injected sensor faults. To validate the findings in real-world AVs, we conduct the physical experiments and the results show the practical significance of our findings.

\thx{\noindent\textbf{Future Work.} 
Based on these findings, we can prioritize which sensor faults require further analysis. We plan to conduct the follow-up research work to \name from the following aspects. (1) Currently, our evaluation focuses on the occurrences of the ADS's safety violations induced by different sensor faults. The evaluation of the severity of the discovered safety violations is orthogonal to the objective of this paper. The further analysis includes detailed assessments of their severity. (2) In further analysis, we will also conduct an empirical study on the robustness of the overall ADS to sensor fault model parameters.}

\section{Data Availability}
The source code of \name is available at \url{https://github.com/ADStesting-test/FADE} or \url{https://zenodo.org/records/15168648}. The experiment results are available at \url{https://zenodo.org/uploads/14015455}.

\section{Acknowledgments}
This work is supported by National Natural Science Foundation of China (62472412, U20A6003), Major Project of ISCAS (ISCAS-ZD-202302) and Project of ISCAS (ISCAS-JCMS-202402). 
This work is also supported by the National Research Foundation, Singapore and DSO National Laboratories under its AI Singapore Programme (AISG Award No: AISG2-GC-2023-008). 
The authors are grateful for the financial support provided by the China Scholarship Council Program (Grant No. 202304910498).

\bibliographystyle{ACM-Reference-Format}
\bibliography{sample-base}










\end{document}